\documentclass{article}

 \usepackage[final]{neurips_2019}

\usepackage[utf8]{inputenc} 
\usepackage[T1]{fontenc}    
\usepackage{textcomp}
\usepackage{algorithm,algcompatible,amsmath}
\usepackage{makecell}
\usepackage{graphicx} 
\usepackage{hyperref}       
\usepackage{url}            
\usepackage{booktabs}       
\usepackage{amsfonts}       
\usepackage{nicefrac}       
\usepackage{microtype}      
\usepackage{amsmath}
\usepackage{natbib}
\usepackage{breqn}
\usepackage{xpatch}
\xpretocmd{\part}{\setcounter{section}{0}}{}{}
\title{Learning to smell for wellness}

\author{%
  Kehinde~Owoeye  \\
Department of Computer Science\\
 University College London\\
  London, WC1E 6EA  \\
  \texttt{ucabowo@ucl.ac.uk} \\
}

\begin{document}

\maketitle

\begin{abstract}
Learning to automatically perceive smell is becoming increasingly important with applications in monitoring the quality of  food and drinks for healthy living. In todays age of proliferation of  internet of things devices, the deployment of electronic nose otherwise known as smell sensors is on the increase for a variety of olfaction applications with the aid of machine learning models. These models are trained to classify food and drink quality into several categories depending on the granularity of interest. However, models trained to smell in one domain rarely perform adequately when used in another domain. In this work, we consider a problem where only few samples are available in the target domain and we are faced with the task of leveraging knowledge from another domain with relatively abundant data to make reliable inference in the target domain. We propose a weakly supervised domain adaptation framework where we demonstrate that by building  multiple models in a mixture of supervised and unsupervised framework, we can generalise effectively from one domain to another. We evaluate our approach on several datasets of beef cuts and quality  collected across different conditions and environments. We empirically show via several experiments that our approach  perform competitively compared to a variety of  baselines.
\end{abstract}

\section{Introduction}
Safeguarding the health and well being of millions of people
in the world most especially in the developing regions of the
world remain one of the seventeen key 2030 Agenda for Sustainable Development of the United Nations~\cite{UN}. To achieve this, the quality of food and drink products remains a target to be  monitored by appropriate authorities to ensure they are safe and healthy for all and sundry. 

\par Due to the proliferation of Internet of things devices, gas sensors  in the form of electronic nose are becoming increasingly available and  important for smelling and tasting chemicals, food and wines~\cite{rodriguez2016electronic,dataset2} for the purpose of assessing their quality. The data obtained from these devices can be used to build a machine learning model with applications in predicting in the future the exact quality of food and drink products at different levels of granularity. However, like most machine learning models, these models on their own do not scale when used in other different but similar domains where the features are different due to covariate shift or when there is a mismatch in the distribution of the labels in the respective domains.

Existing methods proposed towards tackling this problem with respect to time series data have however been designed for domains where  either data has  been collected in well controlled environments~\cite{purushotham2016variational}, simple binary classification problems~\cite{purushotham2016variational} or has considered separately the problem of domain adaptation and semi supervised learning in the same domain with few data points~\cite{zhu2018novel}. There are however problems with these methods in relation to the contexts where they have been used. While weakly supervised learning methods don't scale to other domains, recent work~\cite{saito2019semi} has shown that conventional unsupervised domain adaptation methods designed to produce domain invariant features still perform poorly even when few samples are available in the target domain as well as fail to address adequately classification problems that exists around class boundaries in the target domain. In addition, while domain adaptation problems have hugely focused on generating domain invariant features, in practice, there is a mismatch in the label space coupled with the noisy nature of sensor data.

In this paper,  we consider a scenario where we address the problem of domain adaptation with only few samples (four samples)~\cite{xu2007word} also known as semi-supervised/weakly supervised/few shots domain adaptation. Furthermore, we consider a situation where the classification is more fine-grained with the potential to misclassify for a naive classifier. We propose an approach that leverages a hierarchical  model to find sub-groups in the source domain in an unsupervised manner using clustering. These sub-groups are then trained separately in a supervised learning framework with the aid of a recurrent neural network. A classifier is further trained on four samples per class in the target domain to map these data to the source domain clusters or models where the probability of classifying them accurately is best maximised in addition to training the source domain data in each cluster with the few labeled target domain data. We evaluate our approach on  datasets of beef meat quality of different cuts collected across different spatiotemporal domains. Results on a variety of experiments with these datasets show that our approach  performs competitively compared to competing baselines.

\section{Related Work}
We discuss previous works relative to ours under three broad themes of transfer learning, semi-supervised learning and domain adaptation.

\textbf{Transfer Learning:} Using  all 85 datasets in the UCR archive~\cite{chen2015ucr}, a convolutional neural network (CNN) was proposed to classify time-series~\cite{fawaz2018transfer}. They concluded that source data with some similarity to the target result in positive transfer and negative transfer if there is no similarity. In our case, we are only interested in using dataset with similarities in this case different beef cuts collected across different conditions but with varying difference in the distribution of the input features and labels.

\textbf{Semi-Supervised Learning:} A lot of work have been carried out in this space with respect to time series ~\cite{wei2006semi,guan2007activity}.
One key difference in our approach with respect to these works is that  these methods are only designed for the domain where the source data is collected and perform poorly outside of this domain when the input and target distribution changes.  In addition, we only consider a much more difficult  few shot learning scenario where there are not mot more than four samples per class in the target domain. 

\textbf{Domain adaptation:} Building on the method proposed by~\cite{ganin2016domain}, a variational recurrent adversarial domain adaptation~\cite{purushotham2016variational}  was proposed to generate domain invariant features for healthcare time series data. Compared to the binary classification problem considered in this work, we have focused on even more challenging task of classifying noisy time-series data into four groups where there are non-trivial differences  between the input and label data distribution in the source and target domain. In addition, while they have assumed access to all the input features of the target domain, we assume we only have access to just four samples per class with their associated labels.

\section{Problem formulation \& Training objectives}
\subsection{Problem Formulation}
Given  two time series distributions  $S(x_{t},y_{t})_{t = 1}^{N_{S}}$ and $T(x_{t},y_{t})_{t = 1}^{N_{T}}$ where the former represents the source domain and the latter the target domain while $x_{t}$ and $y_{t}$ represents the input features and the labels at each time-step $t$  respectively. Furthermore, $N_{S}$ and $N_{T}$ may or may not be equal and denotes the respective length of the two distributions and also the two distributions are different but similar in some respects.
We assume during training we have access to all of the data from the source domain and only 4 samples per class from the target domain  $\{S(x_{t},y_{t})_{t = 1}^{N_{S}}, T(x_{t},y_{t})_{t = 1}^{4n} \}$ where $n$ represents the number of unique labels in the target domain distribution. It has been shown that human categorization often asymptotes after just three or four examples~\cite{xu2007word}. Our goal is to build a classifier with almost human level capability  to predict the remaining labels $T(y_{t})_{t = 1}^{N_{T}-4n}$ in the target domain given  $T(x_{t})_{t = 1}^{N_{T}-4n}$.

\subsection{Training objectives}
There are three classifiers in the proposed model each with its training objective (See supplementary material for more details). The overall training objective however is given by:
\begin{dmath}
\underset{\theta_{1},\theta_{2},\theta_{3}}{\mathrm{argmin}} \  E(\theta_{1},\theta_{2},\theta_{3}) = 
\dfrac{1}{N+4n} \sum_{i=1....N+4n} L(y_{i},f(X_{i};\theta_{3})| \\
\dfrac{1}{4n} \sum_{i=1....4n} L(y_{i},f(X_{i};\theta_{2})| \\
\dfrac{1}{N}\sum_{i=1....N}  L(y_{i},f(X_{i};\theta_{1}))))\\
\label{eqn:4}
\end{dmath}

\section{Dataset}
We gathered dataset of  beef meats classified broadly into four groups of excellent, good, acceptable and spoiled with all the datasets skewed towards the spoiled meat. These data have been collected with the aid of electronic nose gas sensors and other sensors to measure variables such as humidity, temperature and TVC (continuous label of microbial population). Each data point in the datasets was recorded per minute in a sequential manner.  \textbf{Dataset 1} is made up of time series data  of beef quality collected across five different instances across two years. Each data instance is 2160 in length~\cite{dataset1}. \textbf{Dataset 2} consist of  an extra-lean fresh beef monitored for about 75 minutes  under fluctuating conditions of humidity and temperature~\cite{dataset2,wijaya2017information,wijaya2016sensor,wijaya2017development}. \textbf{Dataset 3}  contains 12 files of different beef meat cuts such as Inside - Outside, Round, Top Sirloin among others. Eleven gas sensors were used to collect the data~\cite{DVN/XNFVTS_2018}.

\section{Model, Procedures, Experiments \& Baselines}

 We consider experiments (154 in all) across all datasets where we aim to investigate the performance of our model across a variety of contexts. We aim to evaluate the performance of our model when there is  significant difference in the distribution of the input features and labels across the source and target domains.

\subsection{Model architecture}
We use four recurrent neural networks,  LSTM~\cite{hochreiter1997long} overall (two for training the two clusters of the source data alone and another two for training after adding the few target data) with four cells each  and one logistic regression classifier. We use logistic regression to train the few labeled target data with the cluster labels as the input size here is too small for a neural network. The LSTMs with four cells each all employ a many to one classifier with a time-step of two. We Implement the model using Keras and Scikit learn. 

\subsection{Training procedure}
To train the classifier, we use the cluster label where the probability of predicting correctly the label of the data is maximized (two clusters constructed from the source data with the aid of  Gaussian mixture model). In situations where none of the clusters can predict correctly any of the training target data label, we use the cluster label where the frequency of the label is higher.
We run the model ten times settling for the iteration that performed well on the few target data in their new local domain. This model is run a further five times on the unseen target data to find the average classification accuracy. 

\subsection{Baselines}
We use logistic regression (LR), Ada-boost (AB) with a hundred estimators,  Support vector machine (SVM), Semi Supervised learning (SS)~\cite{wei2006semi}, Deep Neural Network (DNN) with two layers each with 256 neurons, Long short term memory (LSTM) with one layer and 4 cells, Recurrent Domain Adaptation Neural Network (RDANN)~\cite{ganin2016domain}. 

\section{Results}
 Results (Table 1), show that our approach outperforms all baselines most of the times and overall across all experiments. Due to lack of the right quantity of data, most of the deep learning models appear to perform poorly.

\begin{table*}[h!]
\centering
\begin{tabular}{c| c c  c c c c c c c}
\hline
 Source-Target   & LR & AB & SVM & SS & DNN & LSTM & R-DANN  & Ours \\
 \hline
$1_{1-5} - 2 $   &  19.59  &  69.43  &  11.26  &  36.24  &  5.95 & 46.02 & 47.18   &  \textbf{79.85} \\

$1_{1}-3_{1-12}$  &  46.58  &  22.00  &  33.60  &  14.65 &  5.30 & 37.55 & 44.86  & \textbf{65.07}   \\

$1_{2}-3_{1-12}$   &  33.97  &  54.73 &  25.89  &  13.59  &  7.97 & 41.21 & 37.61  & \textbf{64.39} \\

$1_{3}-3_{1-12}$   &  36.65  &  54.73  &  26.37  &  13.34  &  3.54 & 60.62 & 33.79  &  \textbf{57.73}  \\

$1_{4}-3_{1-12}$   &  39.41  &  31.37  &  28.42  &  13.63  &  10.53 & 13.31 & 32.62  &\textbf{66.77}  \\

$1_{5}-3_{1-12}$   &  58.99  &  64.53  &  30.02  &  69.32  &  12.24 & 23.74 & 35.43 & \textbf{69.90} \\
 \hline
$2-1_{1-5}$                &  52.06  &  59.44  &  49.03  &  38.81  &  11.69 & 14.08 & 42.77  & \textbf{67.14} \\

$2-3_{1-12}$              &  75.41  &  \textbf{83.82}  &  73.51  &  27.11  &  7.21 & 61.60 & 63.51  &78.59 \\
\hline
$3_{1}-1_{1-5}$         & 54.14  &  46.66 &  45.31  &  \textbf{64.46}  &  8.34 & 49.06 & 38.91  & 52.19 \\

$3_{2}-1_{1-5}$        &   59.34 &  57.83  &  45.09  &  \textbf{59.45}  &  19.44 & 44.27 & 38.91  & 54.85 \\

$3_{3}-1_{1-5}$       &  58.08  &  62.22  &  44.99  &  \textbf{71.04}  &  19.44 & 51.30 & 42.26 & 52.84 \\
$3_{4}-1_{1-5} $      &  \textbf{63.36}  &  62.22  &  47.75  &  47.66  &  14.27 & 43.67 & 39.37  & 62.98 \\

$3_{5}-1_{1-5 }$      &  61.99 &  \textbf{62.22} &  43.88  &  41.04  &  15.58 & 50.44 & 34.33  & 59.89 \\

$3_{6}-1_{1-5}$       &  54.97 &  \textbf{62.22}  &  44.56  &  34.50  &  14.38& 45.37 & 37.59  & 60.23 \\

$3_{7}-1_{1-5} $      & 55.84  &  62.22  &  48.28  &  20.72  &  14.26 & 39.43 & 41.79  & \textbf{65.33} \\

$3_{8}-1_{1-5}$       &  48.93 &  62.22 &  47.35  &  59.46  &  22.07 & 37.93 & 45.19  &  \textbf{62.39} \\

$3_{9}-1_{1-5}$       &   53.40 &  62.22  &  43.48  &  18.96  &  20.17 & 33.31 & 47.12  & \textbf{64.46} \\
$3_{10}-1_{1-5}$     &   58.65 &  62.22  & 47.80  &  \textbf{68.15}  &  15.52 & 47.32 & 30.66  &61.37 \\

$3_{11}-1_{1-5}$     &  55.69  &  62.22  & 43.67  &  \textbf{74.14}  &  13.31 & 48.44 & 35.92  & 56.77 \\

$3_{12}-1_{1-5}$       &  74.21 &  51.11  &  44.01  &  \textbf{100}                &  14.69& 34.20 & 36.69  &  63.46 \\
$3_{1-12}-2$              &  51.56  &  88.08  &  60.28  &  91.33  &  20.52 & 33.60 & 78.74  &\textbf{92.18} \\
\hline
Avg                     &   52.99  &   59.23  &  42.12  &  46.55 &  13.16 &  40.78 & 43.59  & \textbf{66.59} \\
\hline
\end{tabular}
\caption{Classification accuracy all in \%. Our approach can be seen to outperform all baselines most of the time across all experiments suggesting that it is useful when there are both label shifts as well as covariate shifts in the input features.}
\label{tab:4}
\end{table*}

 \section{Conclusion}
 In this paper we have introduced a new approach towards transferring knowledge from one time-series domain to another using only few samples for the purpose of assessing beef quality. Our approach leverages the construction of  unsupervised classification tasks to improve actual beef quality classification tasks. We evaluate our approach on a time series data of  beef quality cuts collected across different conditions. Results across a variety of experiments show that our approach performed competitively compared to competing baselines most especially when the distribution of the target domain labels differs significantly from that of the  source domain. Our work is without its limitations, due to the number of experiments  carried out and the total number of neural networks deployed, we have used the same hyper-parameters across all experiments. Careful tuning of the networks or change of architecture in the future can generate better results. In addition, Just like any other hierarchical model, this approach incurs additional computational cost.
 Furthermore, we envisage distributions with more classes can benefit from deep hierarchical clustering~\cite{heller2005bayesian} compared to the flat clustering we have used. Future work could investigate a combination of some of the techniques used here together with adversarial domain adaptation methods.

\bibliography{references_}
\part*{}

\newpage
\section*{Supplementary Material}
\renewcommand\thesection{\alph{section}}

\section{Approach: Additional Information}

The proposed approach leverages the construction of auxiliary tasks to improve the performance of a downstream supervised learning task. The essence of constructing auxiliary tasks is to aid the efficiency of learning a similar or related task. To construct auxiliary tasks for the purpose of our approach, we aim to find clusters in the source domain data where the probability of classifying each of the few labeled target data is maximized. The task therefore is defined as given the cluster label $C_{k}$ where the probability of classifying the few target labels is maximized, find $ \underset{ \hat{y}}{\mathrm{argmax}}  \ p(\hat{y}|\hat{X},C_{k})$. The choice of the number of clusters is an open question. But it is essential to find a balance between the difficulty of  finding $p(C_{k}| \hat{X}_{i=1...N_{t}})$ and that of $ \underset{ \hat{y}}{\mathrm{argmax}}  \ p(\hat{y}|\hat{X}_{i=1...N_{t}},C_{k})$. 

There are four  benefits of our approach with respect to domain adaptation. First, by finding clusters in the source distribution features, we are able to reduce the mismatch in the distribution of labels. Second, by allocating the target data to source models or cluster where its probability of being predicted is maximised, we reduce the mismatch in the feature distribution between the source and target domain. Third, since sensor data are extremely noisy,  our approach has the potential to ensure extremely noisy inputs are represented in clusters where they appear as outliers enabling the efficient learning of the model parameters. And lastly,  by only using labels for the classes that are far apart in the feature space, it is not necessary in some cases to obtain sample labels of the target data for all the classes as similar input features will be found in the same cluster attached to the same model.

\section{Algorithm}

\begin{algorithm}
     \caption{}
 \begin{algorithmic}[1]
 \STATE\textbf{Input:}  Source data:  $ S(x_{t},y_{t})_{t = 1}^{N_{S}} $,  Target data:  $T(\hat{x}_{t}, \hat{y}_{t} )_{t=1}^{4n}$
 \STATE  \textbf{Output:} Target domain class labels, $\hat{y}^{1},\hat{y}^{2},....,\hat{y}^{N}$
 \STATE  Find clusters $C_{k = 1....k_{n}}$ in the input dataset.
 \STATE Train each cluster  $C_{k}$ with a RNN model $M_{i}$.
 \STATE  Find the cluster $(C_{k})$ / model $(M_{k})$ where $ \underset{ \hat{y}}{\mathrm{argmax}}  \ p(\hat{y}|\hat{X},C_{k})$
 \STATE  Retrain each of the RNN model $M_{k}$ again with the old source data in $C_{k}$ combined with the new data from $T(\hat{x}_{t}, \hat{y}_{t} )_{t=1}^{4n}$
 \STATE Train a classifier to assign target data into the right cluster / model using  $T(\hat{x}_{t})_{t=1}^{4n}$ and labels from $(C_{k})$ .
  \STATE \textbf{for} each datapoint in test data \textbf{do}: 
         \STATE \hskip1.5em Assign data to model $ M_{k} $ from step \textbf{6} using classifier  from step  \textbf{7}.
          \STATE \hskip1.5em  Run the RNN model $M_{k}$ attached to the assigned
          \hskip1.5em cluster $C_{k}$ from step \textbf{6}.
  \STATE\textbf{end} 
  \STATE \textbf{return}  $ \hat{y}_{1},\hat{y}_{2},....,\hat{y}_{N} $
\end{algorithmic}
\end{algorithm}

\section{Training objectives: Additional Information}

There are three classifiers in the proposed model each with its training objective. The training objective for classifying source domain labels alone is given by: 
\begin{equation}
\underset{\theta_{1}}{\mathrm{argmin}}  \ E(\theta_{1}) = \dfrac{1}{N} \sum_{i=1....N}  L(y_{i},f(X_{i};\theta_{1}))\\
\label{eqn:1}
\end{equation}
The training objective for the local domain classification (training few labeled target data with cluster labels) is given by :
\begin{equation}
\underset{\theta_{2}}{\mathrm{argmin}} \ E(\theta_{2})  = \dfrac{1}{4n} \sum_{i=1....4n} L(y_{i},f(X_{i};\theta_{2}))\\
\label{eqn:2}
\end{equation}
Where $n$ is the number of unique classes in the target domain. The loss for classifying labels from all the local domains after adding few labeled data from the target domain (both source and target labels) is given by: 
\begin{equation}
\underset{\theta_{3}}{\mathrm{argmin}} \ E(\theta_{3})  = \dfrac{1}{N+4n}\sum_{i=1....N+4n} L(y_{i},f(X_{i};\theta_{3}))\\
\label{eqn:3}
\end{equation}
The overall training objective is given by  training objective  (equation \ref{eqn:3}) conditioned  on training objective (equation \ref{eqn:2}) which is conditioned on training objective (equation \ref{eqn:1}).
\begin{dmath}
\underset{\theta_{1},\theta_{2},\theta_{3}}{\mathrm{argmin}} \  E(\theta_{1},\theta_{2},\theta_{3}) = 
\dfrac{1}{N+4n} \sum_{i=1....N+4n} L(y_{i},f(X_{i};\theta_{3})| \\
\dfrac{1}{4n} \sum_{i=1....4n} L(y_{i},f(X_{i};\theta_{2})| \\
\dfrac{1}{N}\sum_{i=1....N}  L(y_{i},f(X_{i};\theta_{1}))))\\
\label{eqn:4}
\end{dmath}

\section{Model Architecture}
\begin{figure*}[h]
\centering
\includegraphics[width=12cm, height=5cm]{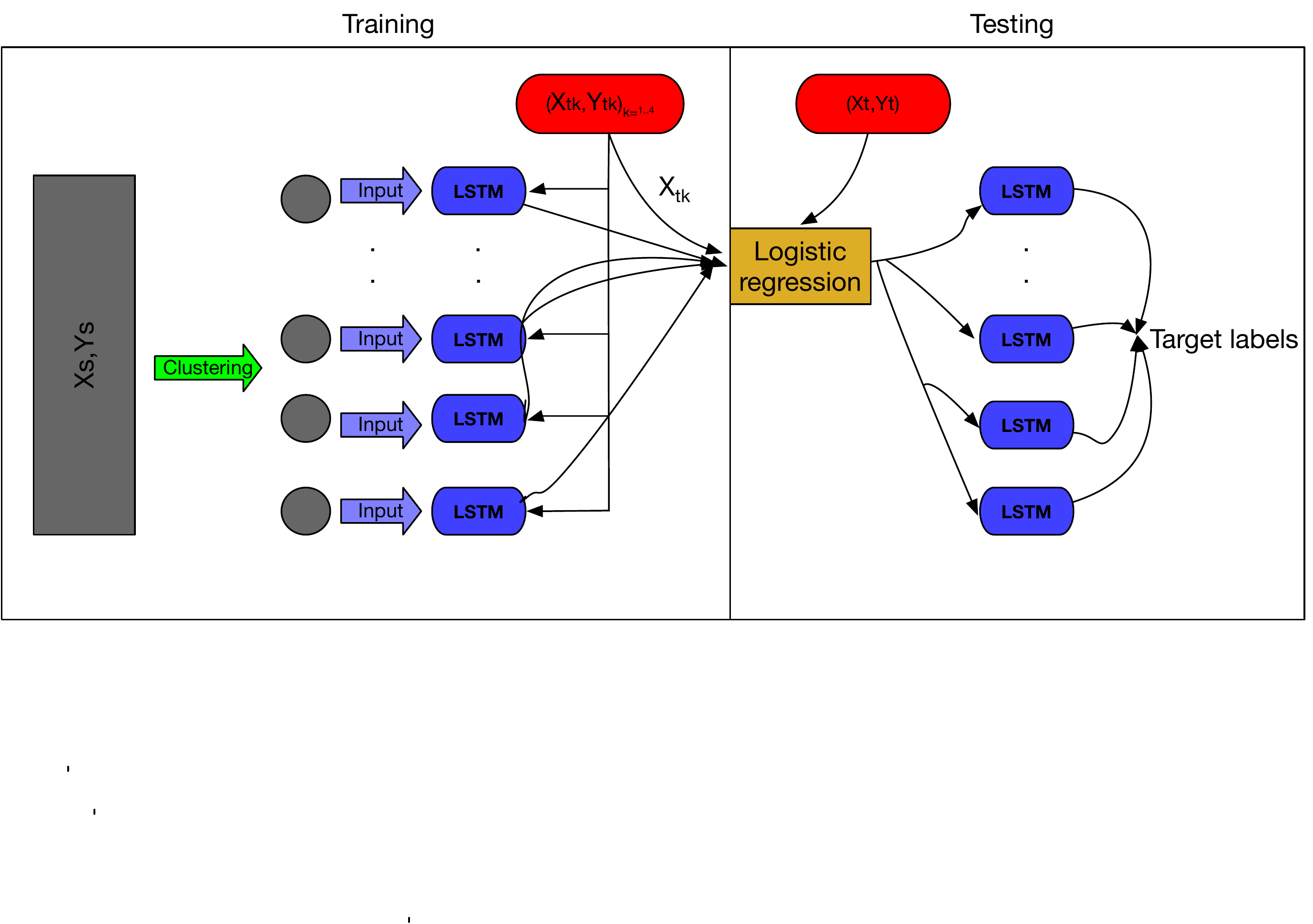} 
\caption{Model architecture showing the structure of our approach.}
\label{fig:1}
\end{figure*}

\section{Baselines: Additional Information}
We compare our approach with several baselines with and without domain adaptation described below. While the domain adaptation baselines are fully unsupervised with the advantage of access to all target features during training, we still aim to compare these methods with our approach to see how these constraints influence performance.

\textbf{Logistic regression (LR)}: We use a multinomial variant with a lbfgs solver. \\

\textbf{Adaboost (AB)}: With 100 estimators.\\

\textbf{Support Vector Machine (SVM):} One versus one. We add the few labeled data from the target distribution to the training data here and also for LR and AB above.\\

\textbf{Semi-Supervised (SS):} Uses one nearest neighbour~\cite{wei2006semi} classifier. We use a variant of this approach where we extend the original approach to a four way classifier since the original approach was proposed for binary classification. We build a  dedicated classifier (assumed to be perfect) for each of the classes containing data from the source domain. We classify each data in the test data by assigning data into class where the one nearest neighbour has the minimum distance with respect to the four classifiers. Test data is added to the training data if the distance to the nearest neighbour is smaller than the mininum distance between samples in the same class in the training dataset. We use only the training data here to assess the ability of  this approach to  generalise when used on datasets from another domain while the test data is added to the training data as discussed above during testing.\\

\textbf{Deep Neural Network (DNN)} : With two layers and 256 neurons each, trained over 100 epochs with dropout = 0.2, softmax layer and adam opimizer~\cite{kingma2014adam}.\\

\textbf{Recurrent Neural Network (LSTM)}: A long short term memory (LSTM) network with 1 layer, timestep = 2, four cells, trained over 100 epochs with dropout = 0.2, softmax layer and adam optimizer~\cite{kingma2014adam}. We use both training data and few labeled target data for training here and also for DNN.\\

\textbf{Recurrent Domain Adaptation Neural Network (R-DANN)}: This is the domain adaptation approach of \cite{ganin2016domain} but with an LSTM in the feature extractor as in ~\cite{purushotham2016variational}. Two layer feed-forward network with 128 neurons  each are further added  to the feature extractor as well as the source and domain classifiers. Relu activation is used throughout the feature extraction network and tanh for the LSTM with a softmax layer for classification.

\section{Data \& Preprocessing: Additional Information}

We provide more information on the datasets we have used here.

We gathered dataset of  beef meats classified broadly into four groups of excellent, good, acceptable and spoiled with all the datasets skewed towards the spoiled meat. These data have been collected with the aid of electronic nose gas sensors and other sensors to measure variables such as humidity, temperature and TVC (continuous label of microbial population). Each data point in the datasets was recorded per minute in a sequential manner.

 \textbf{Dataset 1:} This consists of time series data  of beef quality collected across five different instances across two years. Each data instance is 2160 in length. Nine gas sensors (MQ135, MQ136, MQ2, MQ3, MQ4, MQ5, MQ6, MQ8, MQ9)  were used to collect the data including the humidity and temperature sensors~\cite{dataset1}. 

\textbf{Dataset 2:} This contains an extra-lean fresh beef monitored for about 75 minutes  under fluctuating conditions of humidity and temperature. Ten Gas Sensors (MQ135, MQ136, MQ2, MQ3, MQ4, MQ5, MQ6, MQ7, MQ8, MQ9) were used to collect the data as well as humidity and temperature sensors~\cite{dataset2,wijaya2017information,wijaya2016sensor,wijaya2017development}. 

\textbf{Dataset 3:}  Contains 12 files of different beef meat cuts such as Inside - Outside, Round, Top Sirloin among others. Eleven gas sensors (MQ135, MQ136, MQ137, MQ138, MQ2, MQ3, MQ4, MQ5, MQ6, MQ8, MQ9) were used to collect the data~\cite{DVN/XNFVTS_2018}. 

To ensure the input features are uniform across all datasets collected, we remove the features corresponding to the humidity and temperature variable as well as those corresponding to  the sensors MQ7, MQ138 and MQ137.

\section{Evaluation: Additional Information}
All results are reported using the source data and the few target training data except for SS and R-DANN to demonstrate their inherent limitations in the absence of target domain data. We evaluate all methods on the average classification accuracy from one dataset to another. For example,  $2-1_{1-5}$  means model  trained on dataset 2 containing just one file is tested on dataset 1 with five datasets or files. The average classification accuracy is thus based on the average of the accuracies of the model trained on dataset 2 and tested on the five datasets in dataset 1.

\begin{table}[H]
\centering
\begin{tabular}{c|c|c|c|c|c|c}
\hline
 Dataset& Data length & Beef cut &  \multicolumn{4}{c}{\makecell{Distribution of classes  } } \\
 \hline
 & &  & \makecell{Excellent \\ (Class 1)} & \makecell{Good \\(Class 2)} & \makecell{Acceptable \\(Class 3)} & \makecell{Spoiled \\(Class 4)} \\
\cline{4-7}
 Dataset1 & 2160 & - & 0.111& 0.306 &0.139 &0.444 \\
 &2160 & - & 0.167 & 0.250    &   0.277 & 0.306 \\
 &2160 & - &  0.168 &0.250   &    0.167 & 0.417 \\ 
 &2160 & - & 0.168 &0.250   &    0.194 & 0.389 \\ 
 &2160 & - & 0.168 &0.250   &    0.194 & 0.389 \\ 
 \hline
 Dataset2 & 4453 & Extra-lean & 0.063 &0.046 &0.040& 0.851  \\
 \hline
  Dataset3 & 2220   & Inside \- Outside & 0.0008 & 0.0005 & 0.0005 & 0.998  \\
 &2220   & Round & 0.0008 & 0.0005 & 0.0005 &0.998  \\
 &2220   &  Top Sirloin & 0.135& 0.162 & 0.108 &  0.595 \\  
 &2220   & Tenderloin & 0.135& 0.162 & 0.108 &  0.595 \\  
 &2220   &  Flap meat & 0.135& 0.162 & 0.108 &  0.595 \\  
  &2220   & Striploin & 0.135& 0.162 & 0.108 &  0.595 \\ 
 &2220   & Rib eye & 0.135& 0.162 & 0.108 &  0.595 \\  
 &2220   & Skirt meat &0.135& 0.162 & 0.108 &  0.595  \\
 &2220   & Brisket & 0.135& 0.162 & 0.108 &  0.595  \\
  &2220   & Clod Chuck & 0.135& 0.162 & 0.108 &  0.595 \\
 &2220   & Shin & 0.135& 0.162 & 0.108 &  0.595 \\  
 &2220   & Fat & 0.108 & 0.135 & 0.162 &  0.595 \\
\hline
\end{tabular}
\caption{Distribution of all datasets showing the length, beef cut and the distribution of the different classes of beef quality across three datasets. It can be seen that the distribution of the classes is skewed towards the spoiled meat.}
\label{tab:1}
\end{table}

\end{document}